\documentclass[lettersize,journal]{IEEEtran}
\usepackage{amsmath,amsfonts}
\usepackage{algorithmic}
\usepackage{algorithm}
\usepackage{array}
\usepackage[caption=false,font=normalsize,labelfont=sf,textfont=sf]{subfig}
\usepackage{textcomp}
\usepackage{stfloats}
\usepackage{url}
\usepackage{verbatim}
\usepackage{graphicx}
\usepackage{bbm} 
\usepackage{cite}
\usepackage{color}
\usepackage{xcolor}
\usepackage{booktabs}
\usepackage{multirow}
\usepackage{enumitem}
\usepackage{pifont} 
\newcommand{\cmark}{\ding{51}}%
\newcommand{\xmark}{\ding{55}}%
\usepackage{colortbl} 
\definecolor{LightGray}{gray}{0.95}

\usepackage{hyperref}
\hypersetup{
    colorlinks=true,
    linkcolor=blue, 
    urlcolor=blue,  
    citecolor=purple 
}

\renewcommand{\eqref}[1]{\mbox{Eqn.~\ref{#1}}}

\newcommand{\ie}{\textit{i}.\textit{e}.,~}
\newcommand{\eg}{\textit{e}.\textit{g}.,~}

\newcommand{\gray}[1]{\textcolor{gray}{#1}}

\begin{document}
\title{UNK-VQA: A Dataset and a Probe into the Abstention Ability of Multi-modal Large Models}
%
%
%
%

\author{Yangyang~Guo,~\IEEEmembership{Member,~IEEE,}
        Fangkai~Jiao,
        Zhiqi~Shen,
        Liqiang~Nie,~\IEEEmembership{Senior Member,~IEEE,}
        Mohan~Kankanhalli,~\IEEEmembership{Fellow,~IEEE}
\IEEEcompsocitemizethanks{
\IEEEcompsocthanksitem This research / project is supported by the National Research Foundation, Singapore under its Strategic Capability Research Centres Funding Initiative. Any opinions, findings and conclusions or recommendations expressed in this material are those of the author(s) and do not reflect the views of National Research Foundation, Singapore.
\IEEEcompsocthanksitem Yangyang Guo, Zhiqi Shen, and Mohan Kankanhalli are with
the National University of Singapore, Singapore, 
E-mail: guoyang.eric@gmail.com, dcsshenz@nus.edu.sg, mohan@comp.nus.edu.sg;
\IEEEcompsocthanksitem Fangkai Jiao is with the Nanyang Technological University and I$^2$R, A*STAR, Singapore, Email: jiaofangkai@hotmail.com;
\IEEEcompsocthanksitem Liqiang Nie is with Harbin Institute of Technology (Shenzhen), China. E-mail: nieliqiang@gmail.com.
}}

\markboth{IEEE Transactions on Pattern Analysis and Machine Intelligence}%
{Shell \MakeLowercase{\textit{et al.}}: A Sample Article Using IEEEtran.cls for IEEE Journals}


\maketitle

\begin{abstract}
Teaching Visual Question Answering (VQA) models to refrain from answering unanswerable questions is necessary for building a trustworthy AI system.
Existing studies, though have explored various aspects of VQA but somewhat ignored this particular attribute.
This paper aims to bridge the research gap by contributing a comprehensive dataset, called UNK-VQA.
The dataset is specifically designed to address the challenge of questions that models do not know.
To this end, we first augment the existing data via deliberate perturbations on either the image or question. 
In specific, we carefully ensure that the question-image semantics remain close to the original unperturbed distribution. 
By this means, the identification of unanswerable questions becomes challenging, setting our dataset apart from others that involve mere image replacement.
We then extensively evaluate the zero- and few-shot performance of several emerging multi-modal large models and discover their significant limitations when applied to our dataset. 
Additionally, we also propose a straightforward method to tackle these unanswerable questions.
This dataset, we believe, will serve as a valuable benchmark for enhancing the abstention capability of VQA models, thereby leading to increased trustworthiness of AI systems. 
We have made the \href{https://github.com/guoyang9/UNK-VQA}{dataset} available to facilitate further exploration in this area.
\end{abstract}


\section{Introduction}\label{sec:introduction}
\IEEEPARstart{V}{isual} Question Answering (VQA) serves as a fundamental task in the pursuit of achieving artificial general intelligence.
Conventional approaches often view VQA as a multi-category classification problem, wherein each class aligns with a frequent answer in the dataset under consideration~\cite{VQA2, gqa, vqa-cp}.
These benchmark datasets encompass a spectrum of intricate vision-language comprehension abilities, such as out-of-distribution generalization~\cite{vqa-cp}, compositional reasoning~\cite{clevr}, and utilization of external knowledge~\cite{fvqa, ok-vqa}.

\begin{figure}
  \centering
  \includegraphics[width=1.0\linewidth]{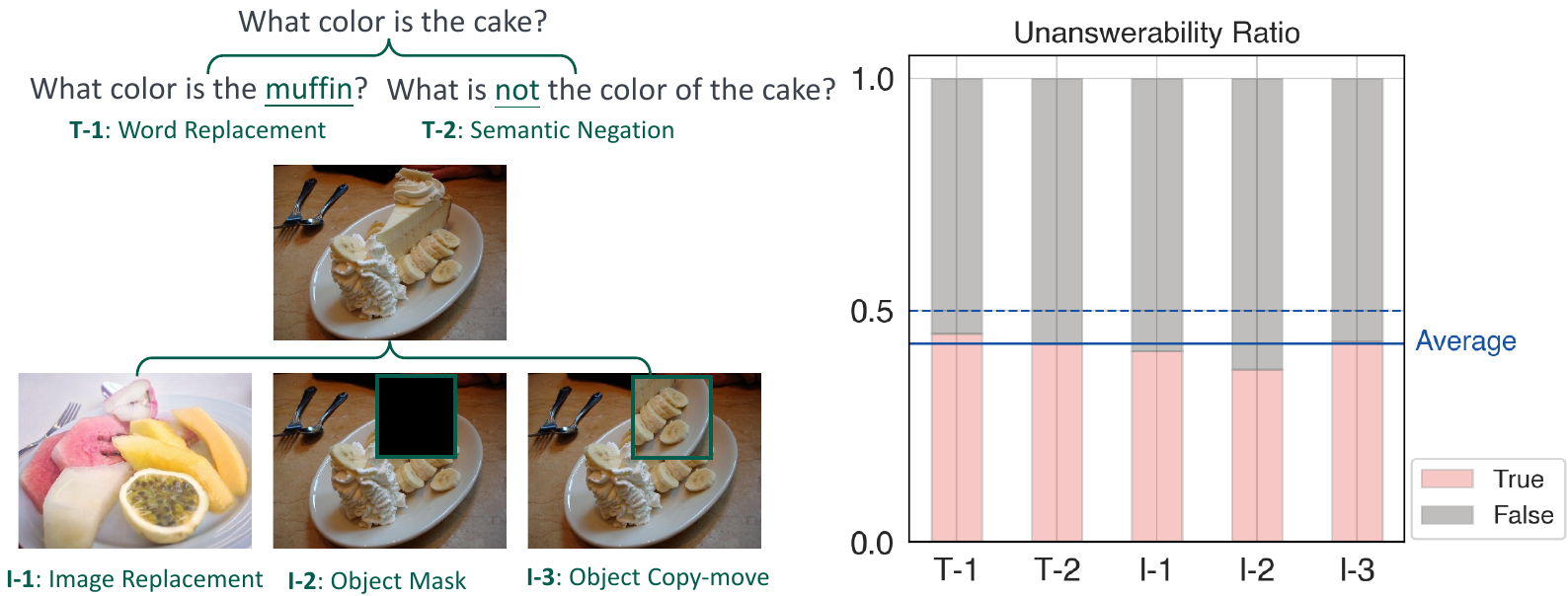}
  \caption{Left subfigure: Five perturbation types from our UNK-VQA dataset and their corresponding exemplars.
  Right subfigure: The unanswerability ratio with respect to each perturbation type.
  }\label{fig:teaser}
\end{figure}

Despite the crucial function of these benchmarks during the early phase of exploration, recent progress has largely been driven by the emergence of large models~\cite{blip-2, gpt-4, flamingo}.
These multi-modal large models overshadow the common training-and-evaluation paradigm in various domains, including VQA.
For example, GPT-4~\cite{gpt-4} achieves a zero-shot accuracy of $\sim$80\% on the VQA v2 dataset~\cite{VQA2}, significantly outperforming most supervised approaches.
Pushing the limits on these traditional well-studied benchmarks thus appears to yield marginal contributions.
While effective, these large foundation models, however, are notorious for being trustworthy.
One demonstration is that they are less capable of abstaining from answering questions that cannot be answered or are beyond their scope of knowledge~\cite{known-llm}.

This paper contributes a dataset that teaches machines to identify and refuse unanswerable visual questions.
Arguably, annotating a large-scale dataset with unanswerable questions is extremely laborious and difficult to control.
To address this challenge, we suggest introducing perturbations to existing data and curating hard instances that can potentially deceive models.
Based on the most popular VQA v2 dataset~\cite{VQA2}, we apply five different types of nuanced perturbations to either the given image or the question, as shown in Fig.~\ref{fig:teaser}. 
To assure high quality, we engage the Amazon Mechanical Turk (MTurk) workers to label them manually.
In short, the total number of participating workers is $>$4,000, and each instance is annotated by a minimum of three workers.
This finally amounts to 10K instances in our dataset, which we name as \textbf{UNK-VQA}.
As can be observed from Fig.~\ref{fig:teaser}, masking the most relevant object (I-2) often leads to more difficulties in performing question answering.
Unlike existing approaches~\cite{abstain-vqa, abstain-vqa-2, realistic-vqa} that employ unsupervised learning techniques to address unanswerable visual questions, our proposed dataset enables model training towards more generalizable VQA methods. 

We evaluate the performance of multiple multi-modal large models including Otter~\cite{otter}, Open-Flamingo~\cite{flamingo}, and InstructBLIP~\cite{instruct-blip}, as well as a proprietary model GPT-4V.
Our findings indicate that:
1) In terms of zero-shot performance, our UNK-VQA dataset reveals limitations in these large models, unlike the impressive results observed in previous general VQA datasets.
2) For few-shot experiments, when we introduce more instances to these models, consistent improvements in their performance can be observed.
Additionally, we also propose a simple method to enhance VQA models with the ability of abstention. 
The key to our method lies in the design of selection functions, which we have developed in several variants, such as a binary classifier and entropy-based methods.
Considering the resource-intensive nature of fine-tuning large models, we apply this method to several conventional and recent medium-sized VL Transformer models.
Furthermore, we also conduct supervised fine-tuning of the LLaVA model~\cite{llava} to observe its performance gain on our UNK-VQA.

This dataset reflects the fact that existing multi-modal large models, in comparison to their counterparts in the language domain, are not as omnipotent.
It sheds light on various issues beyond the challenge of abstaining from answering unanswerable questions. 
Problems such as consistency in VQA~\cite{vqa-consistency}, factuality, and hallucination~\cite{gpt-4} further emphasize the need for future research in this domain.

In summary, we make three contributions in this paper:
\begin{itemize}
    \item We present a new challenging dataset to enable VQA models to abstain from questions that are unanswerable.
    Our dataset helps build models with enhanced trustworthiness and is curated with precise and diverse annotations provided by human annotators.
    \item We extensively study the zero- and few-shot unanswerability capability of several multi-modal large models on the newly introduced dataset.
    \item We introduce a straightforward method for training VQA models that enables them to handle unanswerable questions.
    Moreover, we show the effectiveness of the proposed method as well as the results of supervised fine-tuning of the LLaVA model.
\end{itemize}

\begin{figure*}
  \centering
  \includegraphics[width=0.85\linewidth]{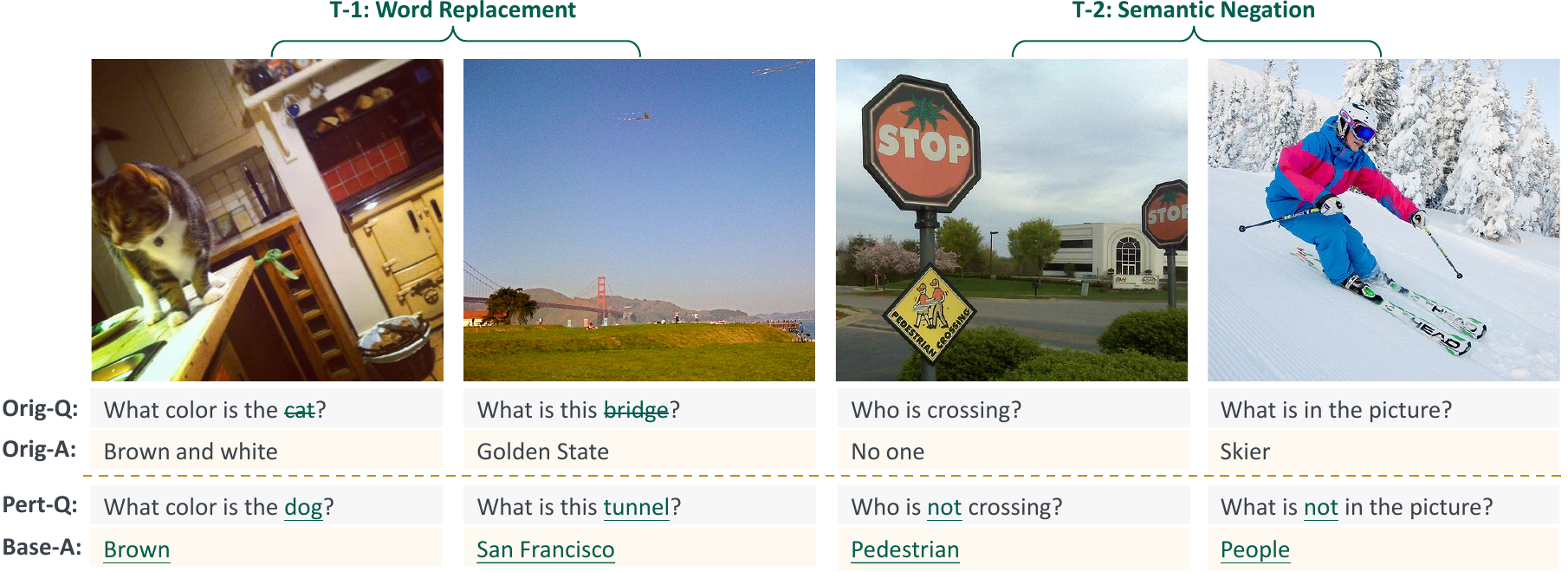}
  \caption{Illustration of text-based perturbations.
  We utilize a strong multi-modal model BLIP~\cite{blip} to generate answers (Base-A) to the perturbed question (Pert-Q).
  The left example shows that when we replace the anchor noun word with an alternative, the baseline still generates a reasonable answer, even though the modified question becomes unanswerable.
  Regarding the semantic negation examples, each question can have an uncountable number of potential answers.
  }\label{fig:text-perturb}
\end{figure*}

\section{UNK-VQA Data Collection and Analysis}
Both traditional approaches and recent multi-modal large models exhibit shortcomings in effectively addressing unanswerable visual questions.
To overcome this, we develop a new benchmark, namely UNK-VQA, that enables both perception and evaluation of the unanswerability capability of VQA models.
Our data collection begins on the basis of two premises:
1) Asking an annotator to write unanswerable questions is highly labor-intensive and uncontrollable.
Specifically, one intuitive approach for humans is to ask irrelevant questions about the given image, which can effortlessly be predicted as \emph{unanswerable}. 
In view of this, we propose introducing perturbations to the existing data.
We then realize that 2) adding perturbations to instances with binary answers (\ie yes and no) can be somewhat useless. 
This is because a question with opposite semantics can often be answered with the other option (\ie yes$\rightarrow$no and vice versa). 
Therefore, we eliminate these instances from our dataset.

\subsection{Perturbation Procedure and Quality Control}
In this work, we introduce five distinct types of perturbations that are applied to either the question or image. 
We provide a detailed explanation of these perturbations below.

\noindent\textbf{Word Replacement (T-1)}
In the current VQA benchmarks, questions typically consist of only a few words, \eg around 10 in length. 
In such cases, the presence of a single salient word, particularly nouns, can have a significant impact on the final answer.
As such, we propose a solution that involves replacing influential nouns with different alternatives. 

For a given question $Q$, we first employ the NLTK toolkit\footnote{https://www.nltk.org/.} to detect nouns and select one noun as the anchor word $w_c$.
After that, we estimate the proximity between $w_c$ and the remaining words in Glove using their pre-trained word embeddings~\cite{pennington-etal-2014-glove}.
The $k$ nearest neighboring words, denoted as $\mathcal{W}_c$, serve as candidates for replacement.
We then enhance the original question by generating augmented questions $\mathcal{Q}_c$. 
This is done by replacing $w_c$ with each $w_i \in \mathcal{W}_c$. 
One instance is shown in Fig.~\ref{fig:text-perturb} that the anchor word \emph{bridge} is replaced with \emph{tunnel}.
We believe that these augmented questions are challenging to answer. 
During this procedure, we eliminate duplicates through lexical rules such as plural and tense comparisons.
Nevertheless, we have observed that some questions with replaced words rarely occur naturally.
This potential shortcut may easily lead the perturbed question to be unanswerable based on semantic coherence alone.
To approach this, we employ a large language model (LM), \ie GPT-2-Large~\cite{radford2019gpt-2}, to filter out examples with high perplexity after augmentation.
The resulting set of questions that pass through the LM filter is defined as:
\begin{equation}
    \begin{aligned}
        \mathcal{Q}_r &= \{Q'|\epsilon \ge \mathrm{LM}(Q')-\mathrm{LM}(Q)\}_{Q' \in \mathcal{Q}_c},
    \end{aligned}
\end{equation}
where
\begin{equation}
    \mathrm{LM}(Q)=-\sum_i^{|Q|}\log p(w_i|w_{<i}, \mathbf{\Theta}),
\end{equation}
where $w_i$ represents the $i$-th token of $Q$, $\mathbf{\Theta}$ is the parameter set of the pre-trained language model, and $\epsilon$ is a pre-defined threshold and we empirically set it to 0.4.

\noindent\textbf{Semantic Negation (T-2)}
Another commonly used approach in text-based perturbation is the semantic negation of the original question.
To ensure the perturbation is effective, we carefully exclude questions that have binary answers to avoid trivial errors.
In practice, we initially perform dependency parsing to identify all verbs and auxiliaries in the question. 
We then add negation to these words, such as converting copulas to their negated form or simply adding \textit{did not} before the verbs.
If none of the above steps are applicable, we check if any negation keywords like \textit{not, hardly,} or \textit{never}, have already been mentioned in the question, and remove them accordingly.
However, it is important to note that using these heuristic rules for negation can sometimes lead to a significant loss in semantic coherence and fluidity.
Similar to the approach used in \emph{word replacement} perturbation, we also rely on a language model to ensure the quality of the synthesized text.

It can be seen from Fig.~\ref{fig:text-perturb} that questions with this perturbation often result in multiple plausible answers.
Due to the imperceptibility of this perturbation to models, the baseline is prone to making errors.

\begin{figure*}
  \centering
  \includegraphics[width=0.95\linewidth]{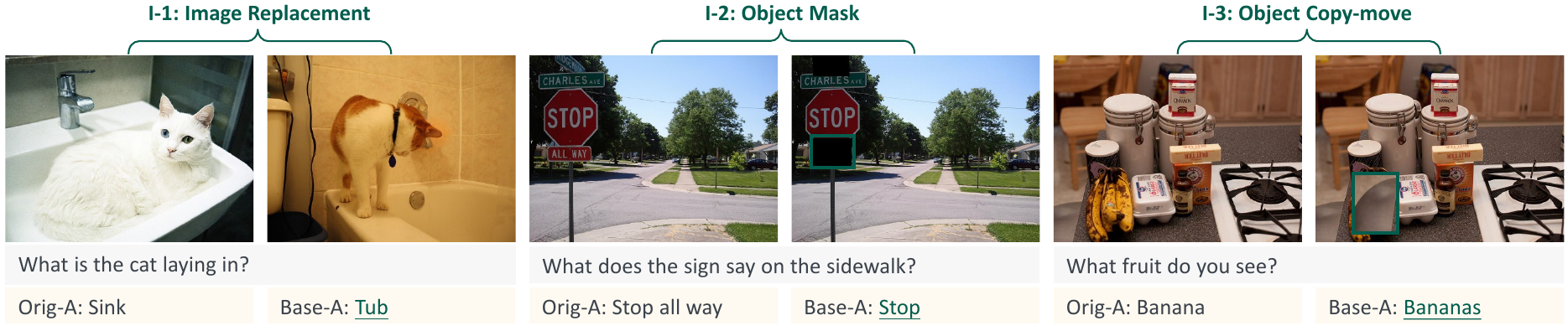}
  \caption{Illustration of three image-based perturbations.
  The BLIP model is also employed to generate answers for the perturbed image (Base-A).
  In the case of image replacement samples, we replace the original image with another image that shares a high degree of semantic similarity.
  For the latter two perturbation types, we cover the most relevant object with a mask and other regions of this image, respectively.
  }\label{fig:image-perturb}
\end{figure*}

\noindent\textbf{Image Replacement (I-1)} 
Unlike the aforementioned text-based perturbations, replacing the given image with another image is a more straightforward task. 
Nevertheless, randomly selecting an image often results in significant semantic drift, making it easy to detect the instance as unanswerable.
In addition, using another image that exhibits similar semantics may lead to the same answer as the original image.
To address both the task difficulty and potential shortcut caused by semantic shifting, our approach for selecting the image candidate is based on two criteria:
1) The candidate image should closely resemble the given image in terms of semantics (such as the \emph{cat} and \emph{bathroom background} in the first example of Fig.~\ref{fig:image-perturb}), ensuring a coherent visual context.
2) The key concepts or objects present in the candidate image and directly related to the question should be excluded, deliberately creating a scenario where the question becomes unanswerable.

To achieve this, we utilize a powerful pre-trained vision encoder, CLIP~\cite{clip}, to extract image features. 
These features are then used to rank all available images based on their similarity to the anchor image, resulting in a selection of the top 50 candidates that share similar semantics with the given image $I$.
Next, we proceed to detect the objects in all the images, as well as the conceptual information within the questions, especially the answers. 
Our objective is to identify and remove images that have a high degree of overlap with the concepts mentioned in the questions and answers. 
To quantify this overlap, we define a semantic overlap score $s_{op}$:
\begin{equation}
    s_{op} = \alpha \cdot \frac{|\mathcal{O}_I\cap\mathcal{O}_{I'}|}{|\mathcal{O}_{I'}|} + \frac{|\mathcal{C}_I\cap\mathcal{O}_{I'}|}{|\mathcal{O}_{I'}|},
\end{equation}
where $\mathcal{O}_I$ and $\mathcal{O}_{I'}$ are the sets of objects in the anchor image $I$ and candidate image $I'$, respectively;
$\mathcal{C}_I$ denotes the concepts (\ie nouns) mentioned in the given question and answer;
and $\alpha$ is a balancing coefficient. 
In this way, the lower the $s_{op}$ value, the more likely the candidate image will be selected as the final replacement image.

\noindent\textbf{Object Mask (I-2) and Object Copy-move (I-3)} 
In addition to perturbations at the image level, we introduce perturbations at a more fine-grained object level. 
Inspired by image manipulation techniques~\cite{manipulation}, we incorporate two popular approaches in this work: object masking and object copy-move.

To achieve this goal, we first extract the concepts mentioned in the given question and answer. 
For each object detected in the image, we compare its corresponding text class with the identified concepts. 
If the object is referenced by the question or answer, we apply either the masking or copy-move approach to this image:
1) For the object masking, we replace the pixel values in the region occupied by the object with 0 (like a mask as shown in Fig.~\ref{fig:image-perturb}).
2) For the object copy-move approach, we randomly select another region in the image that is not relevant (\eg the floor in the last example of Fig.~\ref{fig:image-perturb}), and perform re-scaling and refilling of the previous object region with the newly selected one. 

\begin{table*}[htbp]
    \centering
    \caption{Comparison with several related datasets.
    WP: With Perturbation; CSL: Crowd-source Labeling; TA: Training Apt.} 
    \begin{tabular}{c|c|l|c|r|>{\columncolor{pink!25}}c|>{\columncolor{pink!25}}c|>{\columncolor{pink!25}}c}
    \toprule
    Dataset             & Capability                        & Image Source      & Question Source   & \#Instances   & WP        & CSL       & TA        \\
    \midrule
    FVQA~\cite{fvqa}    & \multirow{2}{*}{Knowledge}        & MSCOCO+ImageNet   & Human             & 5.8K          & \xmark    & \cmark    & \cmark    \\
    OK-VQA~\cite{ok-vqa}&                                   & MSCOCO            & Human             & 14K           & \xmark    & \cmark    & \cmark    \\
    \midrule
    VQA-CP~\cite{vqa-cp}& \multirow{4}{*}{Robustness}       & MSCOCO            & VQA v2            & 658K          & \xmark    & \xmark    & \cmark    \\
    GQA~\cite{gqa}      &                                   & MSCOCO+Flickr     & Automated         & 22M           & \xmark    & \xmark    & \cmark    \\
    AVQA~\cite{avqa-1}  &                                   & Various           & Human             & 123K          & \xmark    & \cmark    & \cmark    \\
    AdVQA~\cite{avqa-2} &                                   & MSCOCO            & Human             & 28K           & \xmark    & \cmark    & \cmark    \\
    \midrule
    VizWiz~\cite{vizwiz}& \multirow{3}{*}{Unanswerability}  & Photo by Blind    & Human             & 31K           & \xmark    & \cmark    & \cmark    \\
    RGQA~\cite{rgqa}    &                                   & GQA testdev       & GQA testdev       & 5.5K          & \cmark    & \xmark    & \xmark    \\
    UNK-VQA (Ours)      &                                   & MSCOCO            & VQA v2            & 10K           & \cmark    & \cmark    & \cmark    \\
    \bottomrule
    \end{tabular}
    \label{tab:dataset}
\end{table*}

\begin{figure}
  \centering
  \includegraphics[width=0.9\linewidth]{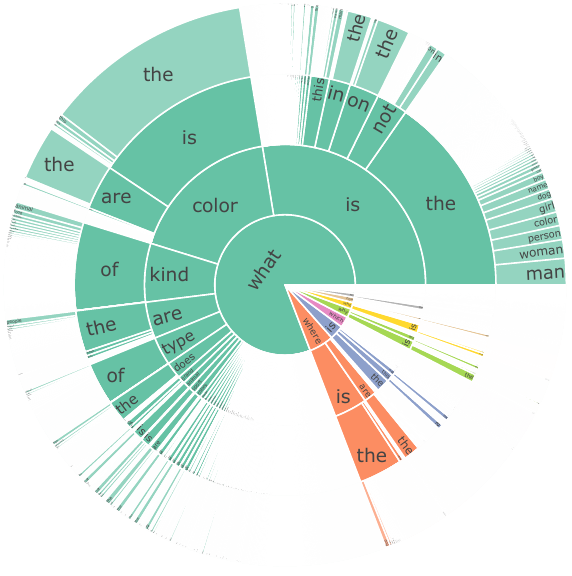}
  \caption{Sunburst distribution of the first four words in the UNK-VQA dataset questions.
  Most questions begin with the word `what'.
  }\label{fig:word-cloud}
\end{figure}

\subsection{Human Labeling} \label{sec: human-label}
After we introduce perturbations to either the given question or image, we engage AMT workers to annotate the results for each instance. 
However, some questions may still be answerable despite these changes. 
To help annotators better understand which questions are unanswerable, we first present an image and a list of unanswerable questions, along with the reasons why they cannot be answered. 
Once annotators have reviewed this information, we then show the image and a related question and ask them to indicate whether the question can be answered correctly.

Arguably, labeling a question with binary answers (yes or no) can potentially result in trivial errors. 
To address this concern, we have implemented a more comprehensive annotation process. 
In addition to labeling whether a question is answerable or not, we require annotators to label multiple following questions.
Specifically, if an annotator determines that a question is unanswerable, we also ask them to provide the reason why it is unanswerable, as well as their unanswerable response to the question. 
The reasons for unanswerability include: \textit{R1) Being unclear to comprehend. R2) Requiring higher-level knowledge. R3) The image lacking important concepts. R4) Having multiple answers}.
The possible responses to unanswerable questions are: \textit{A1) I cannot answer (\eg difficult question). A2) I don't know (\eg beyond my knowledge). A3) Not sure (\eg multiple answers)}.
On the other hand, if annotators select that a question is answerable, we instruct them to indicate which elements (image or question) they believe have been altered. 
Additionally, we provide three answers for such questions: \textit{the original ground-truth answer, a baseline answer offered by~\cite{blip} after perturbation}, 
and \textit{a random answer belonging to the same question type group} (referred to \cite{lan-prior}).
At last, we instruct annotators to label their confidence level on a scale from 1 to 5, where 5 indicates being very confident.

\noindent\textbf{Comparison with related datasets.} 
We compare our UNK-VQA dataset with several representative and related datasets, and the results are summarized in Table~\ref{tab:dataset}. 
It is evident that only RGQA and our UNK-VQA introduce perturbations to VQA instances. 
However, UNK-VQA has two distinct advantages over RGQA:
I) We employ a diverse crowd-sourcing labeling approach, involving over 4,000 annotators, while RGQA only utilized 8 annotators.
II) Unlike RGQA, our dataset allows for training on unanswerable VQA pairs, whereas RGQA is designed solely for evaluation purposes.
Furthermore, it is worth noting that the images in the VizWiz dataset were captured by blind individuals.
Consequently, many images with unanswerable questions exhibit a significant semantic gap between the questions and the visual content, which can result in shortcut predictions by models. 
In contrast, our dataset ensures a strong semantic coherence between questions and images for all instances.
To facilitate training or fine-tuning on UNK-VQA, we further divided the dataset into training, validation, and testing sets, consisting of 7K, 1K, and 2K $\langle$image, question, answer$\rangle$ instances, respectively.

\subsection{Data Analysis}
As depicted in Fig.~\ref{fig:teaser}, object masking tends to result in a higher frequency of unanswerable questions compared to the other four methods. 
This can be attributed to the complete masking of important image regions, which significantly hinders the ability to provide relevant answers.
Additionally, we have conducted further analyses on the collected UNK-VQA dataset, providing valuable insights into the characteristics of the following four questions.

\noindent\textbf{What are the most frequent words in questions?}
We analyzed the first four words of questions in the UNK-VQA dataset and presented the results in Fig.~\ref{fig:word-cloud}. 
Since the dataset excludes binary-answer questions, a large majority of questions begin with the word `what'. 
In particular, `what color' questions are relatively more frequent, likely due to the ease of identifying pertinent objects and keywords associated with them. 
This makes it more effortless to introduce perturbations to either the question words or salient image regions.

\begin{figure}
  \centering
  \includegraphics[width=0.95\linewidth]{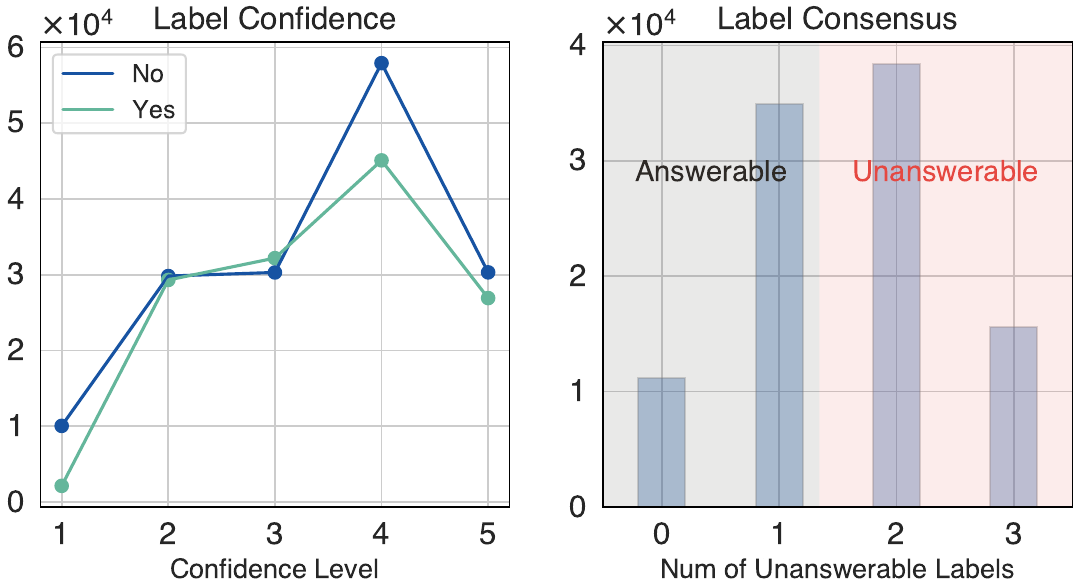}
  \caption{Labeling confidence and consensus among three annotators (y-axis: counted numbers).
  The majority of annotators exhibits a higher level of confidence and can reach a consensus regarding the question answerability.
  }\label{fig:conf}
\end{figure}

\noindent\textbf{Are the unanswerability labels trustful?}
To ensure the accuracy of unanswerability labels, we require annotators to include a confidence level for each survey response. 
As illustrated in Fig.~\ref{fig:conf}, most annotators exhibit a high level of confidence, with a confidence level of 4 being the most frequent. 
Interestingly, for the least confident level (\ie 1), annotators tend to be more confident when labeling a question as \emph{answerable}, indicating that there may be more uncertainty when labeling questions as \emph{unanswerable}. 
To improve the reliability of our labels, we involve at least three annotators for each instance and determine the consensus based on the majority vote rule.

\begin{figure}
  \centering
  \includegraphics[width=0.95\linewidth]{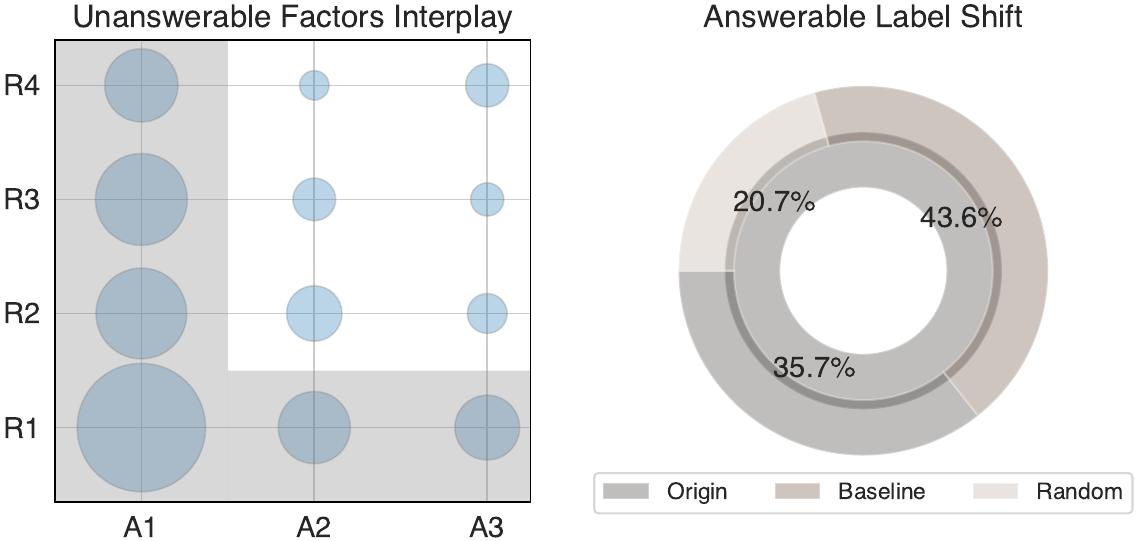}
  \caption{Left: Interplay between answers (A*) and reasons (R*) for unanswerable questions. 
  The most frequent reasons observed are \textit{Being unclear to comprehend}, while the most common answer is \textit{I don't know}.
  Right: Answer shift of answerable questions. 
  Notably, many answers shift from the \textit{Original} ground-truth answers to the \textit{Baseline} predicted answers.
  }\label{fig:answer-interplay}
\end{figure}

\noindent\textbf{What are the most frequent answers and reasons for questions being unanswerable?}
In our annotation process (as described in Sec.~\ref{sec: human-label}), we provide annotators with three answer options and four reasons to select from if they label a question as unanswerable. 
Analyzing the results shown in Fig.~\ref{fig:answer-interplay}, we observe that the most commonly chosen reason for unanswerability is \textit{Being unclear to comprehend}. 
It is notable that even though questions may be unclear to humans, strong baseline models (\eg BLIP~\cite{blip}) often provide answers to these questions with high confidence, raising concerns about the trustworthiness of VQA models.

\noindent\textbf{How do the answers change after perturbation for answerable questions?}
Despite introducing perturbations to either the question or image, there are instances where the answers may still remain unchanged.
In cases where annotators label a question as answerable, we provide three answer options and examine the answer shift as shown in Fig.~\ref{fig:answer-interplay}. 
Notably, 35.7\% of questions still have the same answer as before. 
However, a larger percentage of questions (43.6\%) show a shift towards baseline predicted answers, implying that some answers have changed after perturbations are introduced.

\section{Robustness of Large Models towards Unanswerable Questions} \label{sec: limitation}
Given the remarkable performance of large models across various benchmark datasets, we aim to investigate their effectiveness on our UNK-VQA. 
Specifically, we focus on examining both publicly available models like OpenFlamingo~\cite{flamingo} and a proprietary model, GPT-4V.

\subsection{Evaluation on Open-sourced Models}
\subsubsection{Evaluation Setup}
We pick four multi-modal large models for this evaluation.
\begin{itemize}[leftmargin=1em]
    \item \textbf{Open-Flamingo-MPT-7B} follows similar architecture with Flamingo~\cite{flamingo} but incorporates open-sourced vision and language encoder components, specifically CLIP~\cite{clip} and MPT~\cite{mpt}. 
    The model is trained using a combination of LAION-2B~\cite{laion-5b} and Multimodal C4~\cite{multimodal-c4}, with the weights of both the vision and language encoders frozen.
    \item \textbf{Otter-MPT-7B}~\cite{otter} shares a similar model architecture to that of Open-Flamingo but has been specifically optimized on the Multi-Modal In-Context Instruction Tuning (MIMIC-IT) dataset. 
    This optimization aims to enhance the model's ability to follow the in-context examples accurately for better few-shot inference.
    \item \textbf{Instruct-BLIP-Vicuna-7B}~\cite{instruct-blip} is built on the pre-trained BLIP-2~\cite{blip-2}.
    It leverages a dataset in an instruction-tuning format that has been adapted from public datasets. 
    \item \textbf{Instruct-BLIP-Vicuna-13B} replaces the language model Vicuna-7B with the larger Vicuna-13B~\cite{vicuna}.
\end{itemize}

\begin{table}[t!]
    \centering
    \caption{Zero-shot performance of large multi-modal models on three distinct settings.}
    \begin{tabular}{l|ccc}
    \toprule
    Model                       & BY (\%)           & MC (\%)           & OE (\%)        \\
    \midrule
    Otter-MPT-7B                & 40.87             & 20.33             & 8.70           \\
    Open-Flamingo-MPT-7B        & 31.49             & \textbf{26.52}    & 2.30           \\
    Instruct-BLIP-Vicuna-7B     & 43.98             & 22.38             & 10.86         \\
    Instruct-BLIP-Vicuna-13B    & \textbf{54.82}    & 18.41             & \textbf{12.38} \\
    \bottomrule
    \end{tabular}
    \label{tab:zero-shot}
\end{table}

In addition, we use the accuracy metric for all the experiments and employ the following four evaluation protocols:
\begin{itemize}[leftmargin=1em]
    \item \textbf{Binary (BY)} refers to the binary classification of each question as either \emph{answerable} or \emph{unanswerable}.
    \item \textbf{Multiple-Choice (MC)} involves four answer options for each question in the prompt template. 
    As depicted in Fig.~\ref{fig:answer-interplay}, three options are taken as valid: the \emph{original}, \emph{baseline}, and \emph{random} answers. 
    We add one extra option as \emph{unanswerable}.
    \item \textbf{Open-Ended (OE)} does not include a predefined answer set, allowing for free text generation. 
    \item \textbf{Open-Ended with Hint (OEH)} involves an extra hint that indicates the reason for being unanswerable (we collected in Sect.~\ref{sec: human-label}), compared to OE.
\end{itemize}

For these large models, we employ specific prompts tailored to different settings, enabling the model to directly generate answers. 
The prompts for each setting are provided below\footnote{Certain special tokens specific to particular models, such as \texttt{<GPT>} and \texttt{<User>}, have been omitted in the above examples for illustration.}:

\noindent-- BY
\begin{quote}
\gray{Question: Given the question that \textcolor{darkgray}{\#\textit{question}}, is the above question answerable or unanswerable based on the image?}
\end{quote}

\noindent-- MC
\begin{quote}
\textcolor{darkgray}{\#\textit{question}}

\gray{Options:}
\gray{A.} \textcolor{darkgray}{\#\textit{opt A}}
\gray{B.} \textcolor{darkgray}{\#\textit{opt B}}
\gray{C.} \textcolor{darkgray}{\#\textit{opt C}}
\gray{D.} \textcolor{darkgray}{\#\textit{opt D}}

\gray{The answer is: A/B/C/D.}
\end{quote}

\noindent-- OE

\begin{quote}
\textcolor{darkgray}{\#\textit{question}}
\end{quote}

\noindent-- OEH

\begin{quote}
\textcolor{darkgray}{\#\textit{question}} \gray{If you feel it \#\textit{unanswerable-reason}, you can simply reply ``unanswerable''.}
\end{quote}

Note that in the $k$-shot setting, the above prompts are repeated $k$ times combined with different pre-defined examples, which serve as few-shot prompts. 

\begin{table}[t!]
    \centering
    \caption{OoS (Out-of-Scope) Response Ratio of Otte.
    Generally, we observe that when the model accuracy is higher, the OoS response ratio tends to be lower.}
    \begin{tabular}{c|cc|c|c}
    \toprule
    \#Shots     & \#Ans & \#Una & BY Acc (\%)$\uparrow$ & OoS (\%)$\downarrow$  \\
    \midrule
    0           & 0     & 0     & 40.9                  & 10.4                  \\
    \midrule
    1           & 0     & 1     & 42.0                  & 7.7                   \\
    \midrule
    \multirow{3}{*}{3}
                & 1     & 2     & 13.6                  & 72.1                  \\
                & 0     & 3     & 35.2                  & 23.2                  \\
                & 3     & 0     & 34.6                  & 23.1                  \\
    \midrule
    \multirow{3}{*}{5}
                & 1     & 4     & 7.2                   & 85.3                  \\
                & 0     & 5     & 28.1                  & 39.2                  \\
                & 5     & 0     & 28.3                  & 37.3                  \\
    \bottomrule
    \end{tabular}
    \label{tab:out-of-scope}
\end{table}

\begin{table*}[thbp]
    \centering
    \caption{Effectiveness of the post-hint explanation for addressing unanswerable visual questions.}
    \begin{tabular}{l|c|ccc}
    \toprule
    Model                   & Hint          & 1-shot                                & 3-shot                            & 5-shot                                \\
    \midrule
    \multirow{2}{*}{Otter-MPT-7B}
                            
                            & \xmark        & \textcolor{gray}{10.54}               & \textcolor{gray}{12.17}           & \textcolor{gray}{12.75}               \\
                            & \cmark        &  59.59$_{\textcolor{blue}{+49.05}}$    & 67.39$_{\textcolor{blue}{+55.22}}$& 65.72$_{\textcolor{blue}{+52.97}}$    \\
    \midrule
    \multirow{2}{*}{Open-Flamingo-MPT-7B}
                            & \xmark        & \textcolor{gray}{45.26}               & \textcolor{gray}{54.79}           & \textcolor{gray}{54.57}               \\
                            & \cmark        & 55.05$_{\textcolor{blue}{+9.79}}$     & 55.20$_{\textcolor{blue}{+0.41}}$ & 57.75$_{\textcolor{blue}{+3.18}}$     \\
    \bottomrule
    \end{tabular}
    \label{tab:explanation}
\end{table*}

\subsubsection{Evaluation Results}

\noindent \textbf{Zero-shot Performance.}
We present the zero-shot results in Table~\ref{tab:zero-shot} and observe that Instruct-BLIP-Vicuna-13B outperforms other models in both BY and OE settings. 
Notably, the performance improvements of Instruct-BLIP-Vicuna-13B over Instruct-BLIP-Vicuna-7B highlight the advantages of scaling up the LLM backbone.
Another observation is that Open-Flamingo exhibits superior performance compared to the other three models in the MC setting. 
One possible explanation for this is that, in contrast to BY and OE, MC deviates more from the objective of instruction tuning, \eg open-domain VQA and image captioning. 
Consequently, Open-Flamingo receives less training in instruction tuning, which sets it apart from the other models.

Zero-shot evaluation can have a potential limitation because models may struggle to understand the response format and required knowledge without explicit prompts. 
As a result, instruction-tuned models can generate verbose or simplistic responses, making automatic evaluation challenging.
To address this problem, we further evaluate model performance under the few-shot setting. 

\begin{figure*}
  \centering
  \includegraphics[width=1.0\linewidth]{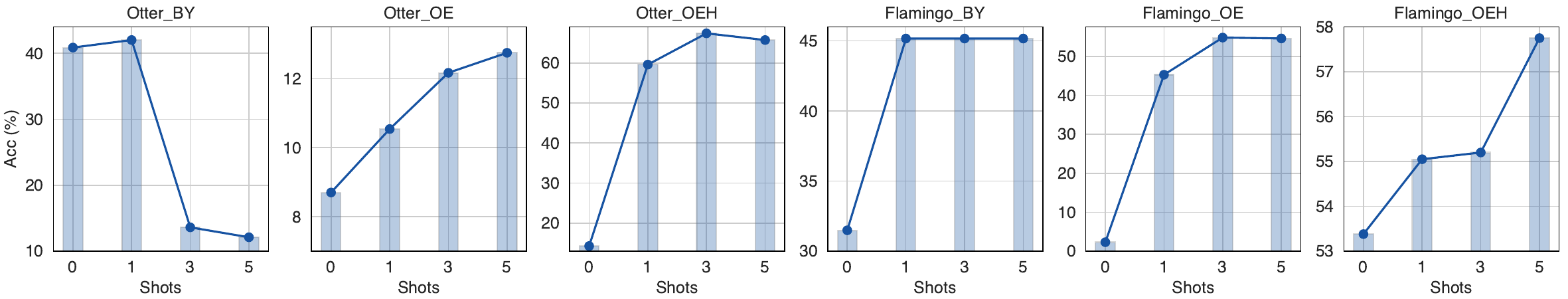}
  \caption{Model performance variance with respect to different number of instance shots.
  }\label{fig:shots}
\end{figure*}

\noindent \textbf{Few-shot Performance.}
We show the results of this experiment in Fig.~\ref{fig:shots}.
One can see that the model performance improves consistently across most experiments.
However, one special case with Otter is that the increased number of shots leads to a decline in performance. 
Upon examining Otter, we found that having more prompted examples often leads to an increase in out-of-scope responses. 
For example, some models tend to respond \textit{I do not know} or attempt to explain the unusual nature of the image.
These are instances where there is no explicit choice of being \emph{answerable} or \emph{unanswerable} in the responses. 
Additionally, the model's performance significantly varies based on the number of answerable and unanswerable examples, as illustrated in Table~\ref{tab:out-of-scope}.

\noindent \textbf{Effectiveness of Post-hint Explanation.}
Note that in Sec.~\ref{sec: human-label}, we ask each annotator to label the unanswerable reason for each instance. 
We further investigate whether the post-hint explanations assist in determining the answerability of visual questions. 
Based on this idea, we provide hints for the unanswerable instances and present the final results in Table~\ref{tab:explanation}. 
The table demonstrates that, in most cases, these post-hints aid the model in comprehending the query more effectively, especially for examples with fewer shots.

\subsection{Evaluation on Proprietary Model}
Initially, we opt to explore the capabilities of the widely used ChatGPT model\footnote{https://openai.com/chatgpt.}.
To adapt the dataset to ChatGPT's text-only requirement, we employ the strong image caption model BLIP-2~\cite{blip-2} to generate detailed captions for each image. 
These captions are then used as the supporting document when asking questions to ChatGPT.
However, the outcomes are far from convincing. 
The generated captions failed to accurately depict the image content, often missing crucial objects or neglecting nouns mentioned in the question. 
Consequently, ChatGPT deemed most questions as `unanswerable', leading to unexpected results.
In light of these limitations, we present to directly employ the GPT-4V\footnote{https://openai.com/research/gpt-4v-system-card.} for our evaluation.

GPT-4V incorporates additional modalities (such as image inputs) into LLMs.
Due to budget constraints, we conducted evaluations on a randomly selected subset of 100 instances from our UNK-VQA dataset.
These instances encompass all five perturbation types.

\noindent \textbf{Results.}
The experimental results of GPT-4V are presented in Table~\ref{tab:gpt-4v}.
There are three notable observations from this table:
\textbf{I}) GPT-4V exhibits significantly superior performance compared to other open-sourced models in the zero-shot setting.
This further proves the superiority of the proprietary GPT-4V over existing open-sourced multi-modal large models.
\textbf{II}) GPT-4V benefits little improvement from multiple-shot learning.
This aligns with recent findings indicating that current LLMs struggle as few-shot information extractors due to task contamination, particularly for proprietary models~\cite{gpt4v-1, gpt4v-2}.
\textbf{III}) There still remains ample room for improvement, underscoring the challenging nature of our UNK-VQA dataset.

\begin{table}[t!]
    \centering
    \caption{GPT-4V results on a subset of UNK-VQA.}
    \begin{tabular}{l|cc|cc|cc}
    \toprule
    \multirow{2}{*}{Model}
            & \multicolumn{2}{c|}{BY (\%)}  & \multicolumn{2}{c|}{MC (\%)}  & \multicolumn{2}{c}{OE (\%)}  \\
            \cmidrule(lr){2-3}              \cmidrule(lr){4-5}              \cmidrule(lr){6-7}
            & 0-shot    & 2-shot            & 0-shot    & 2-shot            & 0-shot    & 2-shot            \\  
    \midrule
    GPT-4V  & 58.0      & 57.0              & 38.0      & 43.0              & 30.0      & 28.0              \\
    \bottomrule
    \end{tabular}
    \label{tab:gpt-4v}
\end{table}

\section{Method}
Existing methods have shown significant limitations in accurately abstaining from unanswerable questions about images. 
In this paper, we propose a straightforward approach that can be easily integrated with existing models to equip them with this capability. 
Due to the computational constraints of training large models, we apply this approach to a selection of conventional models and recent medium-sized pre-trained Transformer models. 
Moreover, we recommend fine-tuning the UNK-VQA data rather than training a VQA model from scratch, thus preserving its original visual reasoning capacity.

\subsection{Preliminary}
The objective of VQA is to generate an accurate answer $\hat{a}$ in response to a given question $Q$ based on an input image $I$. 
This objective can be accomplished through,
\begin{equation}\label{equ:goal}
  \hat{a} = \arg \max_{a \in \mathcal{A}} p(a | Q, I; \mathbf{\Theta}),
\end{equation}
where $\mathbf{\Theta}$ represents the model parameters, and $\mathcal{A}$ denotes the set of all candidate answers. 
The interaction between the question $Q$ and the image $I$ is captured in the feature space $\mathcal{X}$.
Previous approaches have typically treated VQA as a multi-category classification problem, where each possible answer is treated as a distinct class.
Consequently, a standard classifier function $f$ can be defined as $f: \mathcal{X} \mapsto \mathcal{A}$.
In this work, we aim to develop a selective classifier~\cite{abstain-nlp} denoted as $y: \mathcal{X} \mapsto \mathcal{A} \cup \{ \perp \}$, where $\perp$ is a special label that signifies the abstention of prediction. 
Normally, the selective classifier is composed of two functions, $y = (f, g)$, where $g$ is the selective function defined as $g: \mathcal{X} \mapsto \{0, 1\}$.
In this way, given an input feature $x \in \mathcal{X}$, the output of the selective function is determined as follows
\begin{equation}
y(x) = \begin{cases}
    f(x), &\text{if } g(x) = 1, \\
    \perp, &\text{if } g(x) = 0.
\end{cases} 
\end{equation}
Specifically, $g$ is often implemented as a confidence estimator $\hat{g}: \mathcal{X} \mapsto \mathbb{R}$ with a confidence threshold $\theta$,
\begin{equation} \label{equ:theta}
    g(x) = \mathbbm{1} [\hat{g} (x) \geq \theta],
\end{equation}
where $\hat{g}(x)$ indicates the classifier $f$ confidence on the input $x$, and $\theta$ controls the overall prediction versus abstention level.

\subsection{Feature Extraction}
Without loss of generality, we leverage two separate Transformers for the image and question embedding.
Specifically, we employ a ViT encoder~\cite{vit} and a BERT encoder~\cite{bert} to encode the given image $I$ and question $Q$, respectively:
\begin{equation}
    \begin{cases}
        \mathbf{V} &= h_\text{ViT} (I; \mathbf{\Theta}_v), \\
        \mathbf{L} &= h_\text{BERT} (Q; \mathbf{\Theta}_l).
    \end{cases}
\end{equation}
After this, a common approach for modality fusion involves utilizing the features extracted from the [CLS] token of both the image and question encodings,
\begin{equation}
    \mathbf{x} = h_{fn} (\mathbf{V}_{CLS}, \mathbf{I}_{CLS}; \mathbf{\Theta}_{fn}),
\end{equation}
where the fusion operation $h_{fn}$ is not restricted to simple operations such as addition or concatenation; it can also involve another Transformer block. 
Once we obtain the fused feature $\mathbf{x}$, we can easily map it to the answer distribution space using the classifier $f$ and a softmax function.

\subsection{Variants of Selective Functions} \label{sec: selective}
Eqn.~\ref{equ:theta} shows that the selective function is essential to building an answer verification module.
To achieve this, we present two distinct variants to instantiate the selective function.

\noindent \textbf{Classifier-based (CLS)} approach involves the introduction of an additional binary classifier, which facilitates the determination of answerability by comparing the predicted score with a predefined threshold $\theta$. 
Specifically, we re-utilize the fused multi-modal feature $\mathbf{x}$ as inputs to this classifier, represented as $\sigma (\mathbf{W}_b \mathbf{x} + \mathbf{b})$, where $\sigma$ represents the sigmoid function.

\noindent \textbf{Entropy-based (ENT)} variant compares the entropy of the predicted logits with the threshold $\theta$. 
If the entropy of the predicted logits, denoted as $H(\mathbf{\pi}) = -\sum_j^{|\mathcal{A}|} \pi_j \log(\pi_j)$\footnote{In our implementation, we ease the constraint of the infinite range of entropy by comparing the maximum logit with the specified threshold.}, exceeds $\theta$, indicating a higher degree of randomness, the question is classified as unanswerable. 
During training, we assign a uniform distribution $\{\frac{1}{|\mathcal{A}|}, \frac{1}{|\mathcal{A}|}, ...\}$ to label the unanswerable questions, thereby reducing the model's confidence to answer such questions. 
This approach aims to handle cases where the model should refrain from providing a definitive answer.

\begin{table}[t!]
    \centering
    \caption{Model performance on UNK-VQA.
    We did not implement the ENT variant of BLIP because it would have required more computational power due to the generative decoding strategy. 
    Instead, we chose to include an additional label, \emph{unanswerable}, as a generative candidate during both training and inference. 
    }
    \scalebox{0.87}{
    \begin{tabular}{c|>{\centering\arraybackslash}p{1em}>{\centering\arraybackslash}p{1.5em}|ccc|ccc}
    \toprule
    \multirow{2}{*}{Model}                  & \multirow{2}{*}{CLS}  & \multirow{2}{*}{ENT}  & \multicolumn{3}{c|}{valid}               & \multicolumn{3}{c}{test}                  \\
                                                                                            \cmidrule(lr){4-6}                              \cmidrule(lr){7-9}
                                            &                       &                       & Acc$_{b}$     & Acc$_{o}$     & F1$^{W}$      & Acc$_{b}$     & Acc$_{o}$     & F1$^{W}$  \\
    \midrule
    \multirow{2}{*}{UpDn~\cite{bottom-vqa}} & \cmark                &                       & 45.71         & 6.93          & 7.88          & 45.17         & 6.46          & 7.49      \\
                                            &                       & \cmark                & 45.71         & 6.61          & 7.20          & 45.17         & 5.91          & 6.53      \\
    \midrule
    \multirow{2}{*}{LXMERT~\cite{lxmert}}   & \cmark                &                       & 48.47         & 19.64         & 18.47         & 48.50         & 19.45         & 17.36     \\
                                            &                       & \cmark                & 49.04         & 22.75         & 18.68         & 48.40         & 22.40         & 17.68     \\
    \midrule
    \multirow{2}{*}{BLIP~\cite{blip}}       & \cmark                &                       & 58.20         & 40.60         & 36.50         & 57.94         & 40.55         & 36.52      \\
                                            & \xmark                & \xmark                & \textcolor{blue}{45.42}       & \textcolor{blue}{18.23}       & \textcolor{blue}{10.92}                                                                                           & \textcolor{blue}{45.23}       & \textcolor{blue}{18.24}       & \textcolor{blue}{10.94}     \\
    \bottomrule
    \end{tabular}}
    \label{tab:overall}
\end{table}

\section{Experiments}
We conducted extensive experiments on the newly collected dataset spanning two main aspects: 
1) We evaluated the effectiveness of the proposed selective functions;
2) We performed supervised fine-tuning on a strong MMLM - LLaVA~\cite{llava}.

\subsection{Method Evaluation}
\subsubsection{Experimental Settings}

\noindent \textbf{Metrics}. For our evaluation, we adopted three metrics:
\begin{itemize}[leftmargin=1em]
    \item \textbf{Acc$_{b}$} refers to the binary classification accuracy.
    \item \textbf{Acc$_{o}$} compares the overlap between the predicted answer and a pre-defined open set with an \emph{unanswerable} option.
    \item \textbf{Weighted F1 (F1$^{W}$)} is defined as a harmonic mean of the precision P$_j$ and recall R$_j$ ($j$ denotes the $j$-th answer) -- 
    $F1^W_j = 2 \frac{P_j \cdot R_j}{P_j + P_j}$.
    The final metric takes the answer class imbalance into consideration, 
    \begin{equation*}
        F1^W = \frac{1}{\sum_{j \in \mathcal{A}} |\psi_j|} \sum_{j \in \mathcal{A}} |\psi_j| F1^W_j,
    \end{equation*}
    where $\psi_j$ is the ground-truth sample list with answer $j$.
\end{itemize}

\noindent \textbf{Baselines.} We selected three popular baselines to evaluate the effectiveness of the proposed method.
\begin{itemize}[leftmargin=1em]
    \item \textbf{UpDn}~\cite{bottom-vqa} firstly leverages pre-trained object detection frameworks to extract salient object features, enabling high-level visual reasoning. 
    It then employs an attention module to focus on the most relevant objects that are highly associated with the given question. 
    \item \textbf{LXMERT}~\cite{lxmert} is built upon the Transformer encoders. 
    It undergoes pre-training with various pretext tasks on large-scale datasets of image and text pairs, resulting in significant improvement on downstream tasks including VQA.
    \item \textbf{BLIP}~\cite{blip} is a recent strong baseline that is pre-trained for unified vision-language understanding and generation. 
    It effectively utilizes the noisy web data by bootstrapping the captions, where a captioner generates synthetic captions and a filter removes the noisy ones.
    In contrast to the classification-based structure employed by UpDn and LXMERT, we opted to preserve BLIP's original generation objective without making any modifications or adaptations.
\end{itemize}
We applied our proposed method to these baselines to enable them with the ability of abstention from unanswerable questions. 
All the models are maintained with the same experimental settings as the original training protocols.

\subsubsection{Experimental Results}

\begin{figure}[t!]
  \centering
  \includegraphics[width=0.95\linewidth]{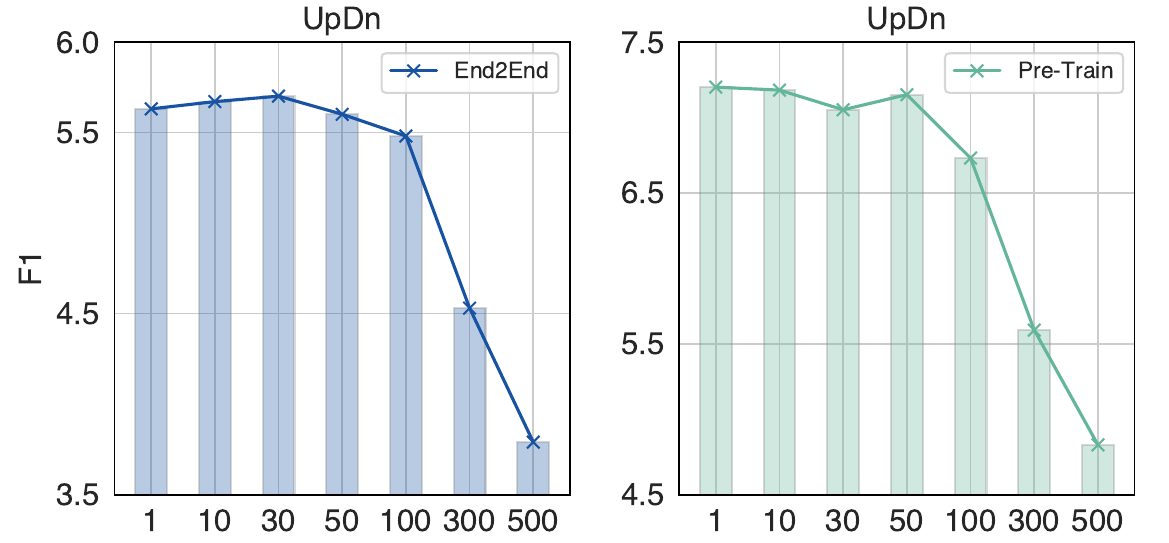}
  \caption{Model performance variance with respect to different threshold values of the entropy-based approach.
  }\label{fig:entropy}
\end{figure}

\noindent \textbf{Overall Comparison.} 
We benchmark three methods, each with two variants, and show the results in Table~\ref{tab:overall}.
The observations are three-fold:
\begin{itemize}[leftmargin=1em]
    \item All the approaches, including the state-of-the-art BLIP model, exhibit relatively lower performance on the UNK-VQA dataset.
    This suggests that these models are less robust to simple perturbations and highlights the need for further improvement on this new dataset.
    \item The performance of the models improves as the model size increases, aligning with recent findings that indicate larger models often endow a better capacity. 
     However, there is an exception with the generation-only version of BLIP (last row), which introduces an \emph{unanswerable} answer choice to the generation pool, making it more challenging compared to other classification-based approaches.
    \item In the case of the vision-language model LXMERT, we found that the entropy-based selective function yields better results on the open accuracy metric, indicating its effectiveness in enhancing model performance.
\end{itemize}

In addition, we also studied the model performance variance with respect to different threshold values on the validation set.
The results are demonstrated in Fig.~\ref{fig:entropy}, which demonstrates that UpDn achieves optimal performance with smaller threshold values, such as 10.
\begin{table*}[htbp]
    \centering
    \caption{Model performance before and after fine-tuning on UNK-VQA.
    All refers to the overall accuracy, while Y/N, Num., and Other represent sub-categories according to different answer types.}
    \begin{tabular}{c|cccc|cccc|cccc}
    \toprule
    Model       & \multicolumn{4}{c|}{UpDn}                         & \multicolumn{4}{c|}{LXMERT}                       & \multicolumn{4}{c}{BLIP}                         \\
                \cmidrule(lr){2-5}                                  \cmidrule(lr){6-9}                                  \cmidrule(lr){10-13}
    FT          & Y/N   & Num.  & Other & All                       & Y/N   & Num.  & Other & All                       & Y/N   & Num.  & Other & All                       \\
    \midrule
    \xmark      & 80.32	& 41.47	& 52.55	& \textcolor{gray}{62.73}   & 87.24	& 53.78	& 61.77	& \textcolor{gray}{71.35}   & 92.56	& 60.58	& 68.30	& \textcolor{gray}{77.41}   \\
    \midrule
    \cmark      & 57.86	& 24.04	& 26.51	& \textcolor{gray}{39.15}   & 78.89	& 40.37	& 51.45	& \textcolor{gray}{61.50}   & 91.33	& 57.50	& 63.78	& \textcolor{gray}{74.41}   \\
    \bottomrule
    \end{tabular}
    \label{tab:degrad}
\end{table*}

\begin{figure*}
  \centering
  \includegraphics[width=0.9\linewidth]{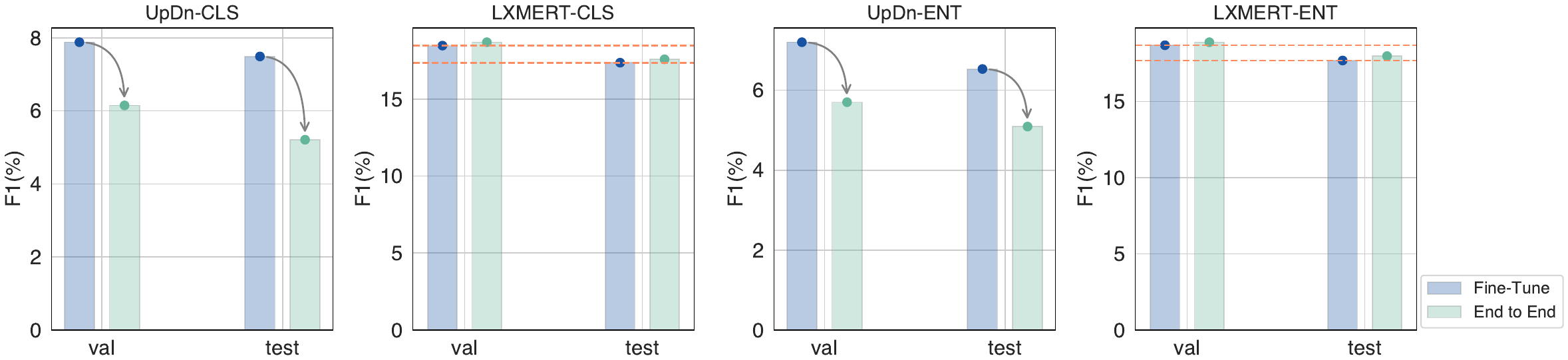}
  \caption{Effectiveness of fine-tuning on UNK compared to that of end-to-end training.
  }\label{fig:finetune}
\end{figure*}

\noindent \textbf{Training from Scratch \textit{v.s.} Fine-tuning.}
In the context of the UNK-VQA dataset, there are generally two training strategies: training a VQA model from scratch and fine-tuning a pre-trained VQA model. 
To compare their performance, we conducted experiments and presented the results in Fig.~\ref{fig:finetune}.
The figure demonstrates that for the UpDn model, fine-tuning outperforms end-to-end training by a significant margin. 
On the other hand, in the case of LXMERT, both training strategies achieve comparable results.
Based on these findings, we advocate for the fine-tuning approach when training a VQA model on the UNK-VQA dataset. 

\noindent \textbf{Re-evaluation on the Original Dataset.}
In addition to the results on UNK-VQA, our focus also lies on evaluating the re-trained model's performance on the original VQA v2 dataset~\cite{VQA2}.
To explore this, we conducted a re-evaluation of the models trained on UNK-VQA and assessed their performance on VQA v2. 
The outcomes of this analysis are presented in Table~\ref{tab:degrad}.
As expected, the model's performance tends to degrade when evaluated on VQA v2 due to the introduction of new instances from UNK-VQA, which serve as out-of-distribution outliers for VQA v2.
Besides this, we have two important findings based on the experiment.
1) Smaller models (UpDn and LXMERT) exhibit a greater degree of performance degradation compared to larger models. 
This finding suggests that smaller models are more susceptible to perturbations and variations in dataset distribution.
2) The generation model, \ie BLIP, shows a relatively lower impact after fine-tuning on UNK-VQA. 
This implies that the generation model is more resilient to the effects of training on UNK-VQA compared to other models.
As such, generating the correct answer, rather than simply classifying it, holds the potential to be more advantageous in future approaches.

\subsection{Supervised Fine-tuning of LLaVA}
LLaVA~\cite{llava} represents a novel end-to-end trained large multi-modal model that combines a vision encoder and an LLM for general-purpose visual and language understanding.
It connects the visual patch-level features from CLIP~\cite{clip} and LLM with a simple linear layer, and is fine-tuned on a multi-modal instruction tuning dataset.
In our experiment, we utilized a more advanced LLaVA-1.6 model and performed supervised fine-tuning on our UNK-VQA dataset.
Specifically, we appended a prompt to each question - \emph{Answer the question using a single word or phrase. 
If you feel you cannot answer this question, simply reply ``Unanswerable''}.
Regarding the LLM, we selected both a weaker Vicuna-13B model and a recent strong Mistral-7B model.

\noindent \textbf{Experimental Results.} 
From the results in Table~\ref{tab:llava}, we can observe that:
\textbf{I}) The LLaVA models exhibit a significant performance advantage over traditional models, as demonstrated in Table~\ref{tab:overall}, owing to their pre-training on more extensive datasets.
\textbf{II}) Both LLaVA models, utilizing different LLMs, show substantial improvement through supervised fine-tuning.
\textbf{III}) Despite its smaller model size, the stronger Mistral-7B LLM notably outperforms the Vicuna-13B model. 
This highlights the pivotal role of LLM capability in advancing multi-modal large models in UNK-VQA.

\begin{table}[t!]
    \centering
    \caption{Supervised fine-tuning results of LLaVA on UNK-VQA.}
    \begin{tabular}{l|c|c|cc}
    \toprule
    Model           & LLM                       & FT        & BY (\%)       & OE (\%)   \\
    \midrule
    \multirow{4}{*}{LLaVA-1.6~\cite{llava}}  
                    & \multirow{2}{*}{Vicuna-13B~\cite{vicuna}}   
                                                & \xmark    & 48.18         & 21.53     \\
                    &                           & \cmark    & 58.00         & 52.49     \\
                    \cmidrule{2-5}
                    & \multirow{2}{*}{Mistral-7B~\cite{mistral}}
                                                & \xmark    & 56.98         & 49.30     \\
                    &                           & \cmark    & 56.35         & 54.16     \\     
    \bottomrule
    \end{tabular}
    \label{tab:llava}
\end{table}

\section{Related Work}
\subsection{Visual Question Answering Datasets}
VQA contributes an essential ingredient to the compelling `AI-Complete' tasks.
In its early emerging stage, significant efforts were dedicated to constructing a general VQA dataset, aiming to encompass all aspects of visual reasoning from textual inputs~\cite{VQA1, tdiuc, visual7w, vg}.
For instance, DAQUAR~\cite{multi-world} and COCO-QA~\cite{coco-qa} employe automated question-answer pair generation techniques, minimizing the need for extensive human labor. 
Visual Madlibs~\cite{madlibs} utilizes fill-in-the-blank templates to generate VQA instances. 
Additionally, benchmark datasets such as~\cite{VQA1, yinyang} collected large-scale human-annotated VQA data, including both natural images and abstract scenes.
With increasing attention to the VQA task, the model performance on these pioneering datasets gradually approached saturation with respect to human performance.

From another point of view, researchers have begun to reconsider more fundamental issues in dataset construction, examining them from a more specific perspective~\cite{vqa-logic}. 
To approach this, some following studies have attempted to introduce additional elements, such as scene text~\cite{scene-text}, multilingual information~\cite{multi-lingual}, and video cues~\cite{videoQA}, to increase the difficulty of answering questions. 
Some other directions like compositional reasoning of questions~\cite{clevr}, and data redistribution based on question types~\cite{vqa-cp} play a crucial role in addressing the notorious bias problem.
In general, reasoning with external knowledge remains a challenging task, while it is effortless for humans. 
To address this, FVQA~\cite{fvqa} and OK-VQA~\cite{ok-vqa} were developed, representing this ability by referring to closed-set supporting facts and open knowledge, respectively. 
Moreover, recent studies have questioned the robustness of existing state-of-the-art VQA models. 
For instance,~\cite{avqa-1, avqa-2} curate adversarial samples to test the robustness of current models when encountering examples in the wild, adopting a human-and-model-in-the-loop procedure to help improve even stronger models.
Furthermore, model explanation has long been a critical problem in the literature. 
Explanatory tools such as post-text generation~\cite{vqa-e, vqa-x} and commonsense reasoning through rationale selection~\cite{vcr} have been used to provide insight into the model's decision-making process.

Pertaining to unanswerable questions, VizWiz~\cite{vizwiz} collects data from blind individuals, often with low-quality photos.
Additionally, there is a significant semantic gap between questions and images.
In contrast, our UNK-VQA provides high-quality images that closely align with the question semantics, which makes recognizing unanswerable questions even more difficult.
Another very recent dataset RGQA~\cite{rgqa} builds unanswerable instances covering the subset categories of our dataset, \ie image replacement and word replacement.
However, the newly collected data do not involve diverse human labeling and RGQA only supports model evaluation. 

\subsection{Unanswerable Machine Reading Comprehension}
Not every question can be answered accurately, which is often overlooked in conventional multiple-choice Machine Reading Comprehension (MRC) datasets that assume all questions have answer options~\cite{uncertainty}. However, in real-world scenarios, it is often necessary to include artificial answer options, such as \emph{none of the above}, to indicate unanswerable questions.
To address this issue, researchers have manually injected unanswerable questions into existing datasets. 
These datasets cover various aspects, including single-round QA~\cite{qa}, multi-round dialogues~\cite{dialog1, dialog2}, more challenging binary questions~\cite{binary-dataset}, and cloze translation~\cite{cloze}.
To tackle the problem of unanswerable questions, most methods incorporate an additional module into the backbone model, such as BERT~\cite{bert}, to abstain from answering when no answer is available for a given question. For instance,~\cite{verify} introduces an auxiliary loss to help verify the legitimacy of the predicted answer. 
NeurQuRI~\cite{inspection} leverages a list of conditions from the question to inspect its answerability. 
Retro-Reader~\cite{retrospective} adopts a retrospective strategy to sequentially perform sketchy and intensive reading to handle such questions.

\subsection{Vision-Language Transformers}
Over the past few years, Transformers have gained significant popularity and have been widely adopted in natural language processing and computer vision~\cite{transformer, bert, vit}. 
With their remarkable performance in these domains, researchers have actively explored the application of Transformers to vision-language tasks.
Specifically, mainstream methods have adopted a \emph{pre-training then fine-tune} paradigm, where models are pre-trained on large-scale datasets~\cite{bert}. 
This approach allows the models to learn general representations from the data before being fine-tuned on task-specific datasets.

Unlike previous pre-training approaches that focused on a single modality, the vision-language domain requires handling two orthogonal inputs. 
The prevalent data format for pre-training in this domain is image-text pairs, where a textual caption is associated with an image. 
Several datasets, such as Conceptual Captions~\cite{cc}, Visual Genome~\cite{vg}, COCO Captions~\cite{coco}, and LAION-400M~\cite{laion}, have been widely adopted for pre-training vision-language models.
The foundation of vision-language Transformers lies in the embedding process of the two modalities. 
For the vision embedding, feature extraction has evolved from grid-based methods~\cite{VQA1}, region-based features~\cite{bottom-vqa} of CNN models, to more recent patch-based features of Transformers~\cite{vl-survey}. 
On the other hand, text tokenization has transitioned from traditional Word2Vec to BERT-style pre-trained embeddings, following the rise of modern language modeling~\cite{transformer, bert}.
In terms of modal fusion, there are generally two approaches: dual-stream and single-stream fusion. 
Dual-stream fusion adopts a late fusion strategy, where the vision and text are separately encoded until a fusion operation is performed to combine the two~\cite{vil-bert, b2t2, lxmert, villa, soho, blip}. 
In contrast, single-stream fusion methods encode the text and vision with a unified Transformer model, where modal fusion is performed beforehand~\cite{video-bert, vl-bert, uniter, oscar, vilt}.
To facilitate training on large-scale captioning datasets, various pretext objectives have been carefully designed, such as masked language modeling~\cite{bert, vil-bert}, masked vision modeling~\cite{uniter, lxmert}, and the image-text matching~\cite{blip, vil-bert}. 

Building upon the achievements of large language models (LLMs), researchers have explored the potential of vision-enhanced LLM foundation models~\cite{fu2023mme, fu2024video, fu2024vita}. 
They have extensively investigated the fundamental capabilities of LLMs, such as instruction tuning, in-context learning, and chain-of-thought, for multi-modal large models. 
Notably, several distinguished models including InstructBLIP~\cite{instruct-blip}, MiniGPT-4~\cite{minigpt4}, and LLaVA~\cite{llava}, have demonstrated the ability to rapidly adapt to new tasks with a few examples.

\section{Conclusion and Future Work}
This paper proposes a novel UNK-VQA dataset to enable VQA models with the ability of abstention. 
We realize that refraining from answering questions that one does not know is a bedrock of intelligence.
Therefore, testing models for this ability is important.
By leveraging this dataset, we probe the robustness of multi-modal large models and identify their relatively less satisfactory performance.
We then propose a straightforward approach to help train VQA models with the inclusion of unanswerable questions.
We evaluate the effectiveness of this approach by integrating it into several baseline methods.
The findings reported in this paper highlight two promising future research directions:

I) Building more challenging benchmarks as the VQA problem is yet far from being solved.
While large models have shown impressive results on conventional datasets, they often lack robustness when faced with small perturbations, as shown in this paper.
To enhance the trustworthiness and generalizability of large models, it is imperative to collect more diverse data and construct more comprehensive benchmarks.

II) Exploring additional possibilities for better multi-modal large model training.
The existing VL large model is still in its infancy, with most efforts focused on aligning them with LLMs.
However, we have discovered that these large models exhibit significant limitations on tasks such as instruction following and visual understanding. 
As such, a more promising direction could involve building multi-modal large models that leverage the intrinsic attributes of multi-modality, rather than relying solely on mapping it back to language.

\bibliographystyle{IEEEtran}
\bibliography{U-VQA}

\begin{IEEEbiography}[{\includegraphics[width=1in,height=1.25in,clip,keepaspectratio]{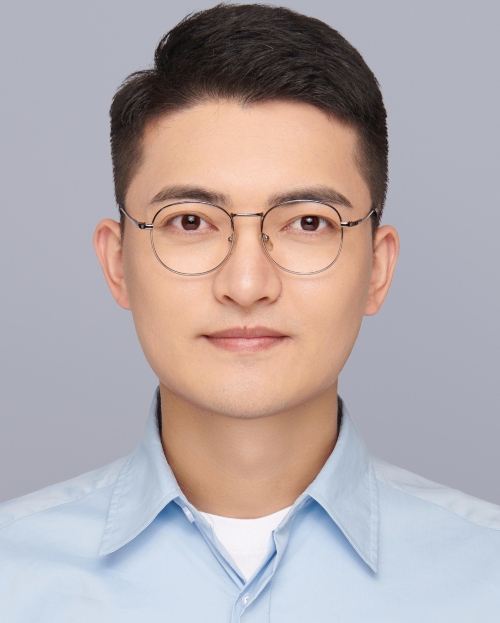}}]{Yangyang Guo}
is currently a research fellow with the National University of Singapore. 
He has authored or co-authored several papers in top journals, such as IEEE TIP, TMM, TKDE, TNNLS, and ACM TOIS. 
He is a Regular Reviewer for journals, including IEEE TIP, TMM, TKDE, TCSVT; ACM TOIS, and ToMM. 
He was the recipient as an outstanding reviewer for IEEE TMM and WSDM 2022.
\end{IEEEbiography}

\begin{IEEEbiography}[{\includegraphics[width=1in,height=1.25in,clip,keepaspectratio]{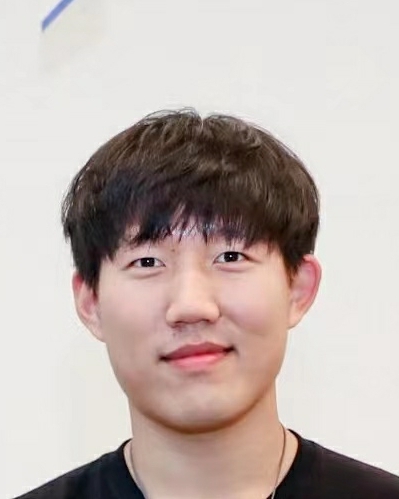}}]{Fangkai Jiao} is currently working toward the Ph.D. degree with Nanyang Technological University, and the Institute for Infocomm Research, Agency for Science, Technology and Research (A*STAR), Singapore. He received the B.Eng. and M.Eng. degrees in software engineering, and computer science and technology from Shandong University, Jinan, China, in 2019 and 2022, respectively. His research interests include self-supervised learning, machine reasoning and large language models.
\end{IEEEbiography}

\begin{IEEEbiography}[{\includegraphics[width=1in,height=1.25in,clip,keepaspectratio]{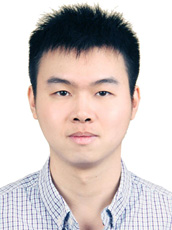}}]{Zhiqi Shen}
is a postdoctoral research fellow with the National University of Singapore (NUS). 
He obtained his Ph.D. degree in computer science from NUS.
He was an intern student at the Institute for Infocomm Research, part of Singapore’s Agency for Science, Technology, and Research from 2014 to 2015. 
His research interest lies in deep learning for computer vision and pattern recognition.
He received the Best Student Paper award at ACM MM 2019.
\end{IEEEbiography}

\begin{IEEEbiography}[{\includegraphics[width=1in,height=1.25in,clip,keepaspectratio]{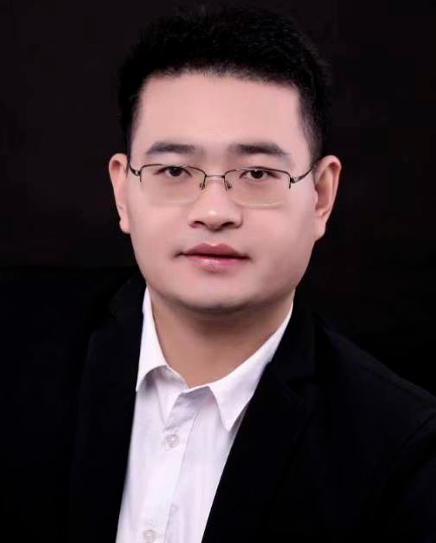}}]{Liqiang Nie}
is currently the dean with the School of Computer Science and Technology, Harbin Institute of Technology (Shenzhen). 
He is a fellow of IAPR and AAIA.
He received his B.Eng. and Ph.D. degrees from Xi'an Jiaotong University and the National University of Singapore, respectively.
His research interests lie primarily in multimedia content analysis and information retrieval. 
He is an AE of IEEE TKDE, IEEE TMM, IEEE TCSVT, ACM ToMM, and Information Science. Meanwhile, he is the regular AC or SPC of ACM MM, NeurIPS, IJCAI, and AAAI. 
He has received many awards, like the ACM MM and SIGIR Best Paper Honorable Mention in 2019, SIGMM Rising Star in 2020, SIGIR Best Student Paper in 2021, and ACM MM Best Paper Award in 2022. 
\end{IEEEbiography}

\begin{IEEEbiography}[{\includegraphics[width=1in,height=1.25in,clip,keepaspectratio]{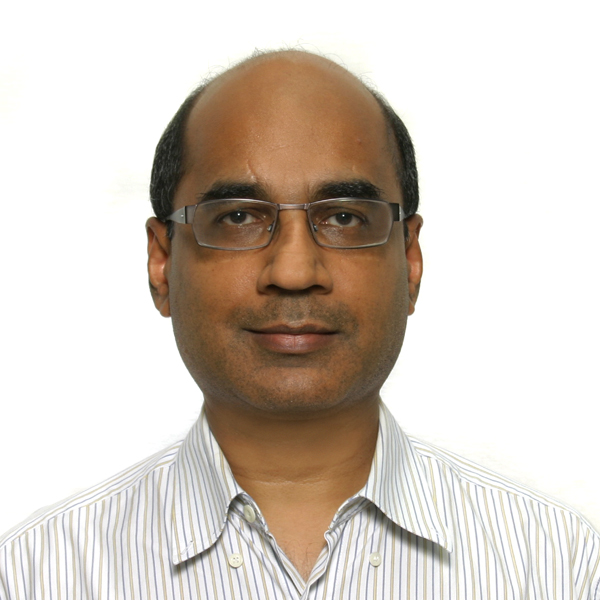}}]{Mohan Kankanhalli}
is the Provost's Chair Professor of Computer Science at the National University of Singapore (NUS) and the Deputy Executive Chairman of AI Singapore. 
He is also the Director of NUS AI Institute, where he leads initiatives on multimodal models and trustworthy machine learning.
Mohan obtained his BTech from IIT Kharagpur and MS \& PhD from the Rensselaer Polytechnic Institute.
Mohan’s research interests are in Multimodal Computing, Computer Vision, and Trustworthy AI. 
His contributions are in image and video understanding, data fusion, visual saliency as well as in content authentication and privacy. 
Mohan is a member of the World Economic Forum's Global Future Council on the Future of Artificial Intelligence. 
\end{IEEEbiography}

\end{document}